%% file: ms.tex
\documentclass[journal,twoside,web]{ieeecolor}
\input{misc/preamble}
\input{misc/commands}
\begin{document}
\title{An Explainable Deep Reinforcement Learning Model for Warfarin Maintenance Dosing Using Policy Distillation and Action Forging}
\author{Sadjad Anzabi Zadeh, W. Nick Street, Barrett W. Thomas
\thanks{
Sadjad Anzabi Zadeh is with Duke Energy Company, NC, 28202 USA (email: sadjad-anzabizadeh@uiowa.edu).
}
\thanks{
W. Nick Street and Barrett W. Thomas are with The University of Iowa Tippie College of Business, IA, 52242 USA (email: nick-street@uiowa.edu, barrett-thomas@uiowa.edu).
}}

\maketitle

\begin{abstract}
\input{00_abstract}

\end{abstract}


\section{Introduction} \label{section: Introduction}
\input{01_introduction}

\section{Background and significance} \label{section: Background and significance}
\input{02_background_and_significance}

\section{Material and Methods} \label{section: Material and Methods}
\input{03_materials_and_method}

\section{Experiments} \label{section: Experiments}
\input{04_experiments}

\section{Results and Discussion} \label{section: Results and discussion}
\input{05_results_and_discussion}

\section{Conclusion} \label{section: Conclusion}
\input{06_conclusion}

\section*{References}

\bibliography{ms.bib}

\end{document}

%% file: misc/preamble.tex
\usepackage{generic}
\usepackage{cite}
\usepackage{amsmath,amssymb,amsfonts}
\usepackage{graphicx}
\usepackage{hyperref}
\hypersetup{hidelinks}
\usepackage{algorithm,algorithmic}
\usepackage{textcomp}
\usepackage{booktabs}
\usepackage[flushleft]{threeparttable}
\usepackage{siunitx}
\usepackage{tabularx}
\usepackage[table]{xcolor}



%% file: 00_abstract.tex
Deep Reinforcement Learning is an effective tool for drug dosing for chronic condition management. However, the final protocol is generally a black box without any justification for its prescribed doses. This paper addresses this issue by proposing an explainable dosing protocol for warfarin using a Proximal Policy Optimization method combined with Policy Distillation. We introduce Action Forging as an effective tool to achieve explainability. Our focus is on the maintenance dosing protocol. Results show that the final model is as easy to understand and deploy as the current dosing protocols and outperforms the baseline dosing algorithms.

%% file: 01_introduction.tex
\IEEEPARstart{F}{inding} the optimal sequence of actions is crucial in many domains, including drug dosing~\cite{anzabizadeh2023optimizing} and online advertisement~\cite{zaho2019deep}. Reinforcement Learning (RL), a broad term for all the methods used to find optimal sequences of actions, has seen dramatic successes in the past decade. Solving protein folding problems~\cite{jumper2021highly} and proposing new algorithms for matrix multiplication~\cite{fawzi2022discovering} are some of the latest examples of successful application of RL in challenging domains. In most cases, deep neural networks learn the optimal action/decision for any situation. Hence they are called Deep Reinforcement Learning (DRL). While conformance to the expected outcome is enough in some domains, critical application domains, such as healthcare, require a deeper understanding of the model's decision-making process. Explainable Reinforcement Learning (XRL) methods attempt to replace or unravel the black box at the heart of DRL models.

This work presents our method of building a dosing protocol for warfarin, an anticoagulant, using XRL. DRL outperforms supervised learning methods regarding dosing performance, known as Percent Time in Therapeutic Range (PTTR) for warfarin~\cite{anzabizadeh2023optimizing}. However, the model is not as available and trustworthy as the current dosing protocols. While in common dosing protocols, the whole protocol is visible as a formula or a table, the proposed DRL protocol was a black box with no visible decision process or logic. It is, of course, a matter of debate whether a nonlinear regression model is interpretable or trustworthy, but the answer is obvious for the DRL model. In this work, we use a Proximal Policy Optimization (PPO) technique to train a deep neural network for maintenance dosing, which determines the change in the current dose to achieve or maintain the therapeutic effect of warfarin. The network receives previous doses and outcomes and returns the percentage change in the dose necessary to achieve and maintain the therapeutic effect for the patient. We apply numerous techniques in the training process to have a dosing protocol that is easier to explain. We coined ``action forging'' for all techniques focusing on the action space. We use the trained DRL model to build a decision tree in the policy distillation phase. The decision tree is then transformed to present the final protocol as a dosing table in a form familiar to practitioners.

The rest of this paper is organized as follows: Section~\ref{section: Background and significance} discusses the need for better warfarin dosing protocols and where this research fits into the current body of knowledge. The problem is explained and formally described as an MDP model in Section~\ref{section: Material and Methods}. We lay out the experiments in Section \ref{section: Experiments} and present the results and their implications in Section~\ref{section: Results and discussion}. We conclude the study in Section~\ref{section: Conclusion}.

%% file: 02_background_and_significance.tex
Warfarin has been used as a common anticoagulant for more than 60 years. While direct-acting oral anticoagulants (DOACs) are starting to replace warfarin, warfarin is still the first line of care in many guidelines~\cite{ho2020trends} and widely used in many parts of the world, such as in Africa~\cite{asiimwe2022ethnic}. Dosing of warfarin is not without its challenges. Since it acts indirectly, patients' diet, lifestyle, and genetic makeup change its effect. The required dose of warfarin can vary from patient to patient over 20 fold~\cite{hamberg2007pk}. Moreover, warfarin has a narrow therapeutic range with severe side effects. Overdosing increases the chance of bleeding, and under-dosing can result in thromboembolism~\cite{shaw2015clinical}. All these features of warfarin make it a perfect candidate for precision medicine~\cite{fusaro2013systems}.

Most dosing protocols for warfarin are developed using clinical trial data and supervised learning methods. International Warfarin Pharmacogenetics Consortium (IWPC) is the biggest of such efforts, with more than 20 collaborating centers and data on more than 6,000 patients. Their thorough use of supervised learning methods, including regression, SVM, and neural networks, shows that the best dosing model to find the maintenance dose is a nonlinear regression model~\cite{international2009estimation}. A \textit{maintenance dose} is a dose that will keep the patient in the therapeutic range for a set number of days. While knowing the maintenance dose is helpful, it will not eliminate the need for a dosing protocol to adjust the dose whenever the blood coagulability goes out of range. Healthcare providers have different dosing protocols. In many cases, these adjustment dosing protocols are tables in which the left column shows possible values of blood coagulability, measured as a dimension-less value called International Normalized Ratio~(INR), and the right column determines the percent of dose change for the patient (see~\cite{ravvaz2017personalized} for an example). In severe cases, such as too high or too low INR values, skipping a dose or a one-time increase in the dose might also be part of the dosing protocol. Besides the change in the dose, these protocols usually determine the next INR measurement interval in days. In severe cases, the measurement and subsequent adjustment might be required as soon as the next day. For less severe cases, however, one week is the most common interval, and if the patient is in the range, the subsequent measurement might be scheduled for up to eight weeks later.

Lately, there has been growing interest in using RL for warfarin dosing. To apply RL, we must define the system's state, denoted by $S$ (what we know about the patient), and the action we take, denoted by $A$ (the dose and duration), which changes the system to a new state ($S'$). During this process, we observe an immediate reward ($R$) which reinforces whether our actions are appropriate. A policy is any mapping from the state space to the action space, and the goal is to find the optimal one. There are many methods to accomplish this, with Q-learning being one of the commonly used. In Q-learning, we aim to learn the value of a state-action pair ($Q(S, A)$), which is the expected cumulative reward if we take action $A$ at state $S$. With $Q$, we can enumerate all possible actions and find the one that maximizes the expected cumulative reward, which is considered the optimal action based on Bellman's optimality principle~\cite{sutton2018reinforcement}.

Zeng et al. (2022) proposed a DRL algorithm based on Q-learning for in-hospital use after surgical valve replacement~ \cite{zeng2022optimizing}. This paper's $S$ comprises 31 demographic, comorbidity, laboratory, and surgery-related variables. The action is the one-day warfarin dose varying from 0~mg to 15~mg. Based on the dataset available to the researchers, they discretized the dose by 0.5~mg steps for quantities less than 6~mg and by 1.0~mg for higher amounts. The current INR and the next-day INR are binned in three groups for the reward function. The reward is a positive value for the change in INR is in the right direction and is negative otherwise. The paper computes the next-day INR using the $k$-Nearest Neighbor method. It predicts the INR of the patient as an average of observations close to the patient in the state space. The results show that their proposed DRL method outperforms clinicians' guidelines in achieving therapeutic range faster, keeping the patients in the therapeutic range longer, and having fewer out-of-range incidents during care and discharge.

Similarly, Q-learning is used in \cite{anzabizadeh2023optimizing} too. However, instead of relying on patient data, they used a pharmacokinetic/ pharmacodynamic (PK/PD) model of warfarin to compute the INR of patients in their study. The PK component models how a medication is absorbed, distributed, metabolized, and excreted, and the PD component describes how drug concentration translates into the effect. The state definition is more straightforward and includes age, CYP2C9 and VKORC1 genotypes, and previous dose and INR information. The dose similarly ranges from 0~mg to 15~mg. and 0.5~mg is the step throughout the range. For the reward, they propose a normalized distance from the mean value of the therapeutic range. While Zeng et al. prescribed the dose for a day and measured INR the next day, Anzabi Zadeh et al. defined the measurement points on days 2, 5, and then every seven days. The results show how the proposed work reacts to high INR values faster and dramatically improves Percent Time in Therapeutic Range (PTTR) compared to commonly used dosing protocols.

Both of the reviewed papers and most of the similar work on the application of RL in drug dosing (See survey in~\cite{yu2019reinforcement}), are black box models that accept the patient state and generate the optimal recommendation. It is not easy to explain such models, and one must check the inner workings of the models and access the logic behind their recommendations in terms of correctness, unbiasedness, and safety, all of which are crucial in the healthcare domain. Despite the progress in making supervised learning models more understandable and interpretable, RL methods are more complex, as they involve long-term consequences, dependency on actions and states, and exploration of the solution space~\cite{rudin2022interpretable}. It is worth noting that explainability and interpretability are just two of several related concepts with contested definitions. As~\cite{barredoarrieta2020explainable} points out, explainability covers techniques that make non-interpretable machine learning models explainable as a post hoc process. According to \cite{rudin2022interpretable}, few methods are available for interpretable RL, such as policy tree representation and relational regression trees, but there are no ``general interpretable well-performing methods'' for DRL. Explainable Reinforcement Learning (XRL) is the alternative that utilizes RL to learn the optimal policy and then attempts to explain the actions post hoc. This paper introduces ``action forging'' techniques to improve the DRL output regarding interpretability. Nevertheless, the proposed techniques are pre-processing techniques for the post hoc explanation of the policy. Therefore, we adopt the term ``explainable'' in this research.

Attempts at interpretable/ explainable RL models are reviewed comprehensively in \cite{glanois2021survey}. In their review, two particular categories are related to what we present in this work. The goal of ``intelligibility-driven regularization'' is to achieve some notion of interpretability by introducing an additional cost function similar to regularization in other areas of machine learning. This approach needs more exploration in RL. In our proposed work, we employ a form of $l1-$regularization to sparsify the action space and discuss our approach to make the action distribution smoother. While these two are similar to the suggested regularizations, we use these techniques to achieve an easier-to-explain policy suitable for post hoc interpretation and not a fully interpretable model.

The other category in~\cite{glanois2021survey} is ``interpretable policy'', in which the goal is to explain the learned policy using an interpretable model. Policy Distillation (PD) matches our proposed work in the methods reviewed in this category. In PD, as introduced in~\cite{rusu2015policy}, the trained model is a teacher whose knowledge will be transferred to another network, the student. Model compression and learning stabilization were PD's main objectives in supervised learning. However, in the context of RL, it has also been used to build interpretable policies using decision trees and soft decision trees~\cite{glanois2021survey}. Similarly, this paper is our attempt to build an XRL model for warfarin dosing. 

%% file: 03_materials_and_method.tex
In this section, we discuss the distinction between initial and maintenance dosing protocols. Then we formulate the problem of maintenance dosing and discuss the solution method. In solving this problem, we introduce ideas and techniques to modify the action space so that the final protocol is easier to explain post hoc. Then we use decision trees to distill an explainable dosing protocol from the final PPO model.

\subsection{Problem Description}
The dosing process starts with prescribing a dose followed by adjusting the dose in the follow-up appointments. Both the initial and adjustment doses depend on patients' response as measured by INR as well as patients' characteristics, including demographic and genetic features (Table~\ref{tab: patient_parameters}). In our previous work~\cite{anzabizadeh2023optimizing}, we proposed a full dosing protocol that covers the whole dosing trial. The common practice, however, is to consider the two dosing phases independently. The initiation phase prescribes the dose. In contrast, the maintenance dose determines the change in the dose. Our focus in this paper is to design a maintenance dosing protocol using RL that is as easy to understand and use as the current dosing protocols while performing better than the current ones. We will use a standard initial dosing protocol (IWPC) to start the dosing process. 


\input{tables/patient_parameters}

\subsection{\texorpdfstring{\uppercase{MDP}}{MDP} Model} \label{paper_2_MDP_model}
We model the maintenance dosing process as a Markov Decision Process~(MDP). Decisions are made in discrete time points and can be one or many days apart. The first decision point $n=1$ occurs on the first day of dose adjustment, and the second decision point is $\tau^1$ days later, where $\tau$ is the number of days and the superscript $1$ indicates the decision point. The sequence of decisions continues by moving forward $\tau^n$ days at each decision point $n$ until we reach the end of the trial on day $T$. The time between two decision points could be part of the decision. However, we fix the intervals in our work to simplify the problem.  

At each decision point $n$, we need the necessary patient information to determine the dose. This information comprises time-invariant components, such as demographic and genetic factors, and time-dependent features, such as the latest INR measurement and the previous dose. We define the state $S^n$ as a tuple including age, CYP2C9 and VKORC1 genotypes, the latest INR measurement~$\mu^n$, the INR value at the previous decision point~$\mu^{n-1}$, the latest dose~$d^{n-1}$ and duration~$\tau^{n-1}$. Note that $d^n$ and $\tau^n$ are the next dose and duration and are not known. As \cite{anzabizadeh2023optimizing} shows, we can include more information by incorporating more of the history of the dosing decisions and INR values. However, the results showed that the improvement from extra information is marginal. Therefore, we include only the most recent history (the equivalent of $h=1$ in that formulation). 

\begin{equation}
\label{eq_2: chap_2 state definition}
    S^n:=\left(Age, CYP2C9, VKORC1, \mu^n, \mu^{n-1}, d^{n-1}, \tau^{n-1}\right)
\end{equation}

Since we want to find a simple, explainable dosing protocol and the results from \cite{anzabizadeh2023optimizing} show that we can exclude genotype information without significant loss of performance, we focus on training a model that relies solely on the time-variant portion of the state definition. To avoid confusion, we denote this definition by $O^n$ since it is the \textit{observed} part of the state.

\begin{equation}
\label{eq_2: chap_2 observation definition}
    O^n:=\left(\mu^n, \mu^{n-1}, d^{n-1}, \tau^{n-1}\right)
\end{equation}

Based on the state/observation of any given patient at decision point $n$, we define the decision $x^n$ as a tuple of $\left(p^n,\tau^n\right)$, where $p$ is the percent change in the dose, and $\tau$ is the pre-specified duration. The next dose to prescribe will be $d^n=d^{n-1}\left(1 + p^n\right)$. The dose can be any value in the range of 0~mg to 15~mg. The practice of any fractional value for the dose is common in dosing protocols, and it is the practitioners' job to determine the actual dose for each day that is practical and close to the prescribed dose. We administer the fractional dose as is and do not modify it for our proposed and baseline protocols. 

The next step is to measure the INR of the patient who received the prescribed dose. In this work, we are not dealing with actual patients for a number of reasons. Ethics of medical trials limits the dose prescription to values that are safe and known to be effective to the patient, which limits RL's exploration of possible solutions. Moreover, DRL algorithms require a large set of training data to converge to the optimal solution, and clinical trials of such magnitude are not feasible. Hence, we use a PK/PD model to simulate patients' reponse to warfarin.

Our MDP model receives the INR value of each patient as exogenous information by running the PK/PD model. The patient's response to warfarin is not deterministic. So, along with the effect of age and genotypes, the PK/PD incorporates noise in concentration levels and INR measurements to account for individual differences. The PK/PD model accepts $\left(Age, CYP2C9, VKORC1, \mu^n, d^n, \tau^n\right)$. It also keeps track of the warfarin concentration from previous doses. The output of the PK/PD is the INR values for $\tau^n$ days for the patient, $W^{n+1}=\left\{ \mu_1^n, \mu_2^n, ..., \mu_\tau^n \right\}$. This vector is the exogenous information in our MDP model. The last INR value is the new observed INR ($\mu^{n+1}=\mu_\tau^n$) that makes the new state $S^{n+1}$. The rest of the values are not observable and cannot be used to make the decision. However, we can use these values to define the reward function and performance metrics.

The reward function provides a scalar value that shows whether the prescribed dose improved or deteriorated the patient's condition. We use daily INR values ($W^{n+1}$) to compute Euclidean distance as a penalty, and the negation of the total penalty is our reward function. That is,
\begin{equation}
\label{eq_2: chap_2 reward function}
    r(S^n, x^n,W^{n+1}) =-c \mathbb{E}\left[\sum_{t=1}^{\tau^n}{\eta^t\left (\mu_m- \mu^n_t\right) ^ 2}\right],
\end{equation}
\noindent
where $\mu_m$  is the midpoint of the therapeutic range ($2.5$ in our case). Parameter $c$ is the normalization factor, and for our therapeutic range of two to three, $c=4$ normalizes the reward so that the reward for both INR values of two and three is $-1.0$. The parameter $\eta$ differentiates this reward function from the one proposed in \cite{anzabizadeh2023optimizing}. That reward function is not sensitive to the direction of change in the INR. If, for example, the INR moves from 2.6 to 2.9 during the period $\tau^n$, it indicates worsening conditions. The reverse trend, however, shows that the dose is moving the INR in the right direction. In the case of the previous reward function, these two scenarios would produce the same reward and will not guide the model to discern good and bad dosing decisions. In the new reward function, $\eta$ is the amplifying factor; a number slightly greater than one that penalizes moving away from $\mu_m$. As we move away from the dosing day, $\eta^t$ increases, putting more weight on the later INR differences. Note that in this formulation, we need the exogenous information revealed after making the decision to compute the reward. For this reason, we have $W^{n+1}$ rather than $W^n$.

We define the objective function as maximizing the total reward for all patients for the duration of the experiment. That is:
\begin{equation}
    F^\star=\max_{\pi \in \Pi} \mathbb{E}\Big[\sum_{P\in \Phi}{\sum_{n}{r^\pi(S_P^n, x^{\pi,n}, W_P^{n+1})}\Big|S_P^0}\Big],
\end{equation}
\noindent
where $\Pi$ is the set of all possible policies, and $\Phi$ is the set of all patients.

\subsection{Proximal Policy Optimization}
To solve the dosing problem described in the previous section, the dosing trial starts with one of the most commonly used dosing algorithms, IWPC~\cite{johnson2017clinical}. Then we use Proximal Policy Optimization (PPO) to find the optimal maintenance dosing protocol. PPO is a policy gradient algorithm, which means the actions are learned directly, as opposed to value function approximation methods such as Q-learning~\cite{sutton2018reinforcement}, in which the algorithm learns to assign value to state-action pairs and then finds the action with maximum value. Since we want to manipulate actions to generate the explainable dosing policy, policy gradient algorithms are better options for XRL.

Implementation-wise, two separate neural networks are trained in the PPO algorithm: the actor, which receives the state and provides the probability distribution over all actions, and the critic, which learns the value of each state ($V(s)$). The critic is used during the training to mitigate the high variance of the observed reward~\cite{schulman2017proximal}. A mini-batch of observations is fed to the model in each training iteration, and the actor and critic are trained independently. This algorithm is an on-policy method since the observations result from using the same PPO model to produce the actions.

\begin{figure}[!ht]
    \centering
    \includegraphics[width=\columnwidth]{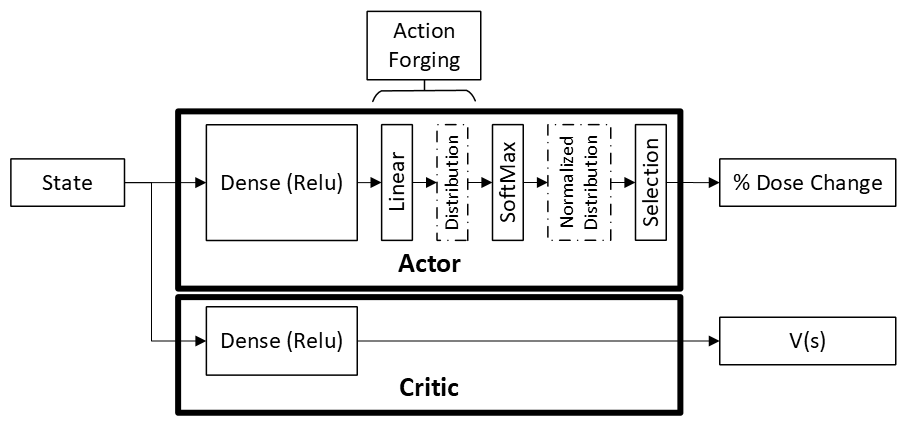}
    \caption{The PPO structure with Action Forging}
    \label{fig: ac_w_action_forging}
\end{figure}

Figure~\ref{fig: ac_w_action_forging} shows our implementation of the PPO model. The actor consists of a set of fully-connected layers with ReLu activation functions and one layer with linear activations. To pick an action, the output of the linear layer passes through the softmax function to be normalized and the action is selected. During the training, the selection is random according to the action distribution. In inference, we deterministically pick the action with the highest probability. The critic has a simpler architecture of fully-connected layers with ReLu activation functions. The output of the critic, $V(s)$, is used during the training only to help with the learning process.



\subsection{Action Forging}
Moving towards explainability, we implement several action-forging techniques. By action forging, we mean tools and ways to change the shape of the action distribution based on our understanding of the distribution and our goal of achieving an explainable policy. The main premises in this section are:
\begin{itemize}
    \item \textit{Fewer dose changes are preferred}. Reducing dose change frequency reduces adverse events and improves patients' medication compliance and effectiveness~\cite{shi2007impact}.
    \item \textit{Fewer dose change options are preferred}. Considering the established protocols of dose adjustment, which are mostly in tabular form, we want the practitioner to see a smaller table of possible decisions to avoid mistakes and improve the usability of the protocol. Also, in the post hoc analysis, improved explainability can compensate for an acceptable and marginal loss in performance.
\end{itemize}

All ideas and techniques in this section are applied to the linear layer and its output in the actor network~(Figure \ref{fig: ac_w_action_forging}). In other words, we change the un-normalized action probabilities, not the normalized ones.

\paragraph{Action Regularizer}

To have fewer actions, we introduce an action regularizer. We discretize the action space, aiming to produce a dosing protocol similar to the ones in practice. When changing the dose, most protocols depict the logic as a table with actions as pre-determined percent changes. Hence, we also discretize the percent dose changes. However, deciding which percentages to include in the action space is challenging. A bigger action space means more freedom for the model to choose the appropriate action. Nevertheless, it also translates into fine-tuning the prescribed dose by the model and increasing the number of dose changes in the course of treatment. A smaller action space, on the other hand, will allow us to extract clear-cut rules, but can negatively impact the performance if the values in the action space are not determined properly.

To let the model choose which actions work best as part of the training process, we penalize the model's number of actions available to the model. This regularization eliminates some actions from the action space. To enforce this, we ensure the probability of those actions as an output of the neural network is zero, irrespective of the input, which can be achieved by forcing the weights of the respective output neurons to zero. Since we want weights to be zero, the $l_2$ norm of the weights of the output neurons corresponding to the eliminated actions should be zero. Since we want to reduce the number of actions, the penalty is an $l_1$ regularizer. Therefore, the action regularizer is

\begin{equation}
    L:=\left\| \left\|W_j \right\|_2 \forall j\in \{1...\left| \Omega \right|\} \right\|_1
\end{equation}

\noindent where $\Omega$ is the action space and $W_j$ is the vector of the weights and bias of the neuron corresponding to the $j$th action.

One challenge with the proposed regularizer is that an algorithm like PPO does not generally include a mechanism for exploration, such as $\epsilon$--greedy, other than relying on the action distribution. As a result, it will always explore an action that has yet to be eliminated. When the regularizer eliminates an action, the corresponding neuron will always generate zero as its output. If an action is not selected, there will be no gradient for it, and without the gradient, its probability will not change in backpropagation. In short, an eliminated action will never become active again. So, having a small regularizer coefficient is imperative to ensure enough exploration and avoid early elimination of actions.

\paragraph{Action Focus}

A dosing policy comprises a logic to choose a dose and a duration. The logic depends on the state, and the result of the logic can be any of the permissible actions. Sometimes, the dose change is inevitable since failure to change the dose will place the patient out of the therapeutic range for INR. In other instances, dose change is either the result of the protocol being too eager to keep the INR close to the middle point of the therapeutic range or the indifference between adjacent dose values. We want to limit the number of dose changes to the greatest extent possible. Hence, we promote the action of ``no dose change'' ($0\%$ dose change). To this end, we can increase the probability of this action over other actions while allowing the model to change the dose if necessary. So, we need to ``focus'' on $0\%$ dose change by increasing the probability of that action compared to small dose changes and not the whole action space. A wavelet function $\psi()$ can provide this focus. Note that ``a wavelet is a transient waveform of finite length'' \cite{dondurur2018acquisition} with applications in wavelet analysis. In this work, we only use a wavelet to modify the action probabilities, not as used in wavelet analysis.

During the early stages of training, applying the wavelet will affect exploration and damage the model's performance. Therefore, we need a parameter to determine the shape of the function. Early in the training, the function should be a flat line to have no effect. Gradually the function should take its complete form and full effect. We define this as a two-argument function:

\begin{equation}
    \Psi(\omega, \delta)=\omega + h(\delta)\psi(\omega)
\end{equation}

\noindent
where $\omega$ is the vector of action probabilities, $\delta$ is the training step, $h$ is a non-decreasing function of $\mathbb{R}\rightarrow[0, 1]$ that determines how much effect to expect from the wavelet function, and $\psi$ is the wavelet function. 





\begin{figure}[!ht]
    \centering
    \includegraphics[width=\columnwidth]{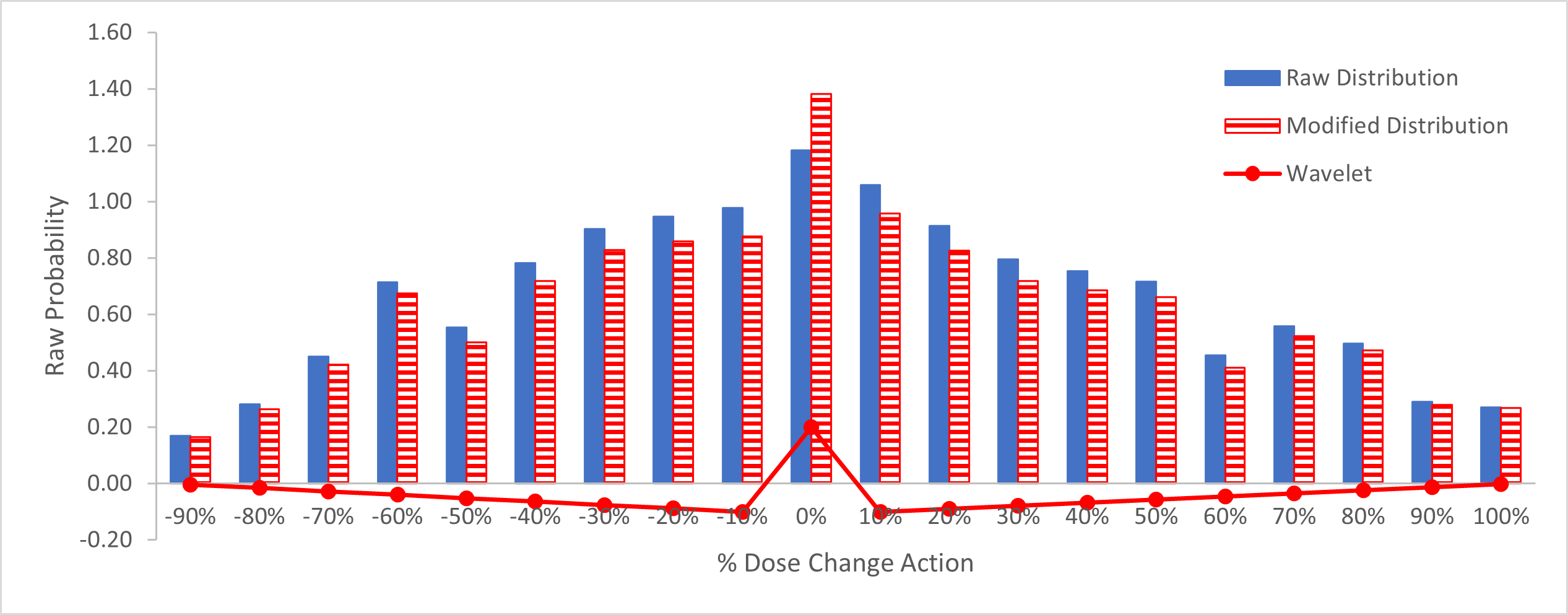}
    \caption{The effect of applying a wavelet function to the action distribution}
    \label{fig: action_focus}
\end{figure}

The wavelet we used is a piece-wise linear function with a positive value at $0.0\%$ action and zero or negative for the rest of the actions.

\begin{equation}
\label{eq_2: pointyhat_wavelet}
    \psi(t)=\left\{
        \begin{array}{ll}
           u  & \mbox{if } x = 0 \\
          \frac{-d}{r}\left|x\right| + d & \mbox{otherwise} \\
        \end{array}
    \right.
\end{equation}

\noindent
where $u$ is the value we want to add to the probability of no dose change action, $r$ is the maximum percent change, and $d$ determines the level of decreasing the neighboring actions probabilities. Figure~\ref{fig: action_focus} shows the shape of the function for $u=0.2$, $d=-0.1$, and $r=1.0$ (red line) and how it changes the probabilities of the actions (solid orange bars) to emphasize more the zero dose change action (striped red bars). When the model eventually learns to keep the patients in the therapeutic range, this increase in the probability of the zero dose change action persuades the model to pick that action more than its neighbors. As a result, this action becomes preferred over its close neighbors and will be the dominant action during the inference.

%% file: tables/patient_parameters.tex
\begin{table}[ht!]
    \caption{Characteristics of the virtual patients (adopted from~\cite{anzabizadeh2023optimizing})}
    \label{tab: patient_parameters}
    \centering
    \begin{threeparttable}
        \begin{tabular}{l @{}S[table-format=3.2, table-column-width = 4cm]@{}}
            \toprule
            Characteristic  & {mean$\pm$SD} \\
            \midrule
            $Age$ (yr)\tnote{1} & 67.3$\pm$14.43 \\
            Weight (lb)\tnote{2} & 199.24$\pm$54.71 \\
            Height (in)\tnote{3} & 66.78$\pm$4.31 \\
            \rowcolor{lightgray} Sex (\%) & \\
            ~~~Female   & 53.14 \\
            ~~~Male     & 46.86 \\
            \rowcolor{lightgray} Race (\%) & \\
            ~~~White     & 95.18 \\
            ~~~Black     &  ~4.25 \\
            ~~~Asian     &  ~0.39 \\
            ~~~American Indian/ Alaskan     &  0.18 \\
            ~~~Pacific Islander     &  0.0001 \\
            \rowcolor{lightgray} Tobacco (\%) & \\
            ~~~No      & 90.33 \\
            ~~~Yes     & 9.66  \\
            \rowcolor{lightgray} Amiodarone (\%) & \\
            ~~~No      & 88.45 \\
            ~~~Yes     & 11.54  \\
            \rowcolor{lightgray} Fluvastatin (\%) & \\
            ~~~No      & 99.97 \\
            ~~~Yes     & 0.03  \\
            \rowcolor{lightgray} $CYP2C9$ (\%) & \\
            ~~~*1/*1     & 67.39 \\
            ~~~*1/*2     & 14.86 \\
            ~~~*1/*3     &  9.25 \\
            ~~~*2/*2     &  6.51 \\
            ~~~*2/*3     &  1.97 \\
            ~~~*3/*3\tnote{4}     &  0.00 \\
            \rowcolor{lightgray} $VKORC1$  & \\
            ~~~G/G       & 38.37 \\
            ~~~G/A       & 44.18 \\
            ~~~A/A       & 17.45 \\
            \bottomrule
        \end{tabular}
        \begin{tablenotes}
            \item[1, 2, 3] {Age, weight, and height are clipped to the ranges of [18, 100], [70, 500], and [45, 85], respectively, based on a dataset of 10,000 virtual patients provided by~\cite{ravvaz2017personalized}.}
            \item[4] In the implementation, we assumed the probability of observing this genotype to be $2.0\times 10^{-4}$.
        \end{tablenotes}
    \end{threeparttable}
\end{table}

%% file: 04_experiments.tex
The 90-day dosing trial starts with the initial dosing protocol, IWPC, which determines the dose for the first four days. The maintenance dosing protocol is responsible for the dosing every seven days starting day 5. Some protocols include an ``adjustment protocol'' between the initial and maintenance dosing. In most dosing protocols, the maintenance dosing period (post initial or post adjustment) starts later than our proposed work (day 6 or 8 in \cite{ravvaz2017personalized}).

All trained models share a similar architecture. The actor is a four-layer fully connected network with ReLu activation functions in hidden layers and linear activation functions in the last layer. A softmax function is applied to the output of the layer to normalize the probabilities of the actions. The critic network also has four layers but with a single output. Table~\ref{tab: model_parameters} shows all the parameters.

\input{tables/model_parameters}

The action forging techniques have their hyper-parameters, and they can interact with each other and impact the training. We did not do a comprehensive study to find the optimal combination. Instead, we start by finding the best coefficient for the regularizer using a grid search over its coefficient. Then, we introduce the wavelet function and adjust its parameters. For $h$ function, that controls the effect of wavelet, in Equation~\ref{eq_2: pointyhat_wavelet}, we use a two-parameter sigmoid function:

\begin{equation}
    h(\delta)=\frac{1}{1 + e^{-c_1\left(\delta-c_2\right)}}
\end{equation}
\noindent
where $c_1$ is the scaling parameter and $c_2$ is the translation parameter. We set these parameters in our experiments to $c_1=1e-3$ and $c_2=50$.

During the training, patients are generated randomly. Patient characteristics (Table~\ref{tab: patient_parameters}) show an imbalance in the distribution of CYP2C9. To account for this imbalance, we set the minimum probability of each variant during the training to $0.1$. In other words, variants ``*1/*1'' and ``*1/*2'' will have prevalence values of $45.14\%$ (down from $67.39$) and $14.86\%$ (same as before), and the rest of the variants will have $10\%$ chance of expression. At each pass, the model interacts with 500 patients, collects the dosing trajectories to build the mini-batch of observations, and uses it to train. The Percent Time in Therapeutic Range (PTTR) is the performance measure we track to ensure the training is improving. It shows the percentage of time patients in the mini-batch were in the therapeutic range (INR of two to three). Training starts with a warm-up period and then stops if the model does not see an improvement for a set number of training passes.

The results are compared with two baseline protocols. Similar to our model's setup, we start the baseline dosing protocols with IWPC protocol. Both baselines utilize Lenzini adjustment protocol~\cite{lenzini2010integration}, which includes previous INR values as well as genetic information to adjust the dose. In the maintenance phase, Aurora uses ``Aurora best-practice standard dose warfarin therapy protocol''~\cite{ravvaz2017personalized} and Intermountain makes decisions based on ``INR-based Intermountain Healthcare Chronic Anticoagulation Clinic Protocol''~\cite{anderson2007randomized}. Both maintenance dosing protocols rely on INR as the only factor to determine the percent change necessary for the patient.

After training, testing, and comparing our proposed model with the baselines, we build the explainable protocol. We rely on Python's sci-kit-learn package's implemented decision tree algorithm and default parameters. Like baseline maintenance dosing protocols, we focus on the INR as the only covariate to decide the dose. So, the final protocol will be a table with INR ranges on the left side and the percentage of dose change on the right side.

%% file: tables/model_parameters.tex
\begin{table}[!ht]
    \caption{Experiment setup}
    \label{tab: model_parameters}
    \centering
    \begin{tabular}{l l}
        \toprule
        Parameter & Value \\
        \midrule
        \rowcolor{lightgray} Actor & \\
        ~~~layers & $256, 256, 128, 64$ \\
        ~~~learning rate function & exponential decay (with staircase) \\
        ~~~initial learning rate & $1e-4$ \\
        ~~~learning rate steps  & $1,000$ \\
        ~~~learning rate decay  & $0.8$ \\
        ~~~training iterations  & $20$ \\
        ~~~GAE lambda           & $0.97$ \\
        ~~~target KL divergence & $0.02$ \\
        ~~~entropy loss coefficient   & $0.0$ \\

        \rowcolor{lightgray} Critic & \\
        ~~~layers & $256, 256, 128, 64$ \\
        ~~~learning rate function & exponential decay (with staircase) \\
        ~~~initial learning rate & $1e-5$ \\
        ~~~learning rate steps & $1,000$ \\
        ~~~learning rate decay & $0.8$ \\
        ~~~training iterations & $80$ \\

        \rowcolor{lightgray} Other Parameters & \\
        ~~~reward clip  & ($-30$, None) \\
        ~~~clip ratio   & $0.2$ \\
        ~~~discount factor & $0.5$ \\

        ~~~dosing duration & $90$ days \\
        ~~~initial dosing & IWPC \\
        ~~~initial dosing duration & $4$ days \\

        ~~~buffer size & $500$ observations \\
        ~~~warm-up period & $20,000$ virtual patients \\
        ~~~test size & $2,000$ virtual patients \\

        \bottomrule
    \end{tabular}
\end{table}

%% file: 05_results_and_discussion.tex
Table~\ref{tab: pttr_base_vs_baselines} presents the PTTR values of the base dosing protocol (limited state definition $O$), the full state definition ($S$), and two baseline protocols. The base model achieves an average of 84.5\% PTTR (standard deviation $0.09$ percentage points), significantly better than the baseline results. The most significant gap is in the highly sensitive category where the best baseline achieves the PTTR of 25.9\%, almost a third of what our proposed model can achieve. In all sensitivity levels, the standard deviation of our proposed model is smaller than the baselines, which indicates that more patients receive adequate care under the proposed protocol. The full-state model is not significantly different from the base model, which suggests that the recent dosing history (the previous and current INR and the previous dosing decision) provides enough information for the model to make a sound decision.


\input{tables/pttr_base_vs_baselines}

Table~\ref{tab: base_vs_regularized} shows how the regularizer changes the number of actions and the performance of the models. The base model allows all 21 possible dose change percentages. Dealing with the test set of 2,000 patients, the model uses seven actions and achieves 84.5\% PTTR. Note that in the test phase, we pick the action deterministically by picking the action with the highest probability. Therefore, even though all 21 actions are available, only seven are used, indicating that we only need some possible actions to perform well. The regularizer shows its effect gradually. For small regularizer coefficients, the number of actions used in the test set decreases. 
Then the number of available actions, which is the main focus of the regularizer, starts to diminish at higher coefficients. The minimum happens at $0.1$, where there are four actions left. More robust regularization reverses the process and increases the number of available actions. This behavior is because a strong regularizer does not allow for any of the actions to maintain a high probability. As the regularizer pushes down all action probabilities, they all remain comparable. As a result, all actions will have a chance to be selected as the dosing decision. The minimum action space, which belongs to the regularizer coefficient value of $0.1$, comprises $\{-60\%, -10\%, 50\%, 90\%\}$. 

\input{tables/base_vs_regularized}

In the next step, we apply the action modifier and the regularizer. We used the piece-wise linear function in Equation~\ref{eq_2: pointyhat_wavelet} with parameters depicted in Figure~\ref{fig: action_focus} along with the sigmoid function with parameters $c_1=10^{-3}$ and $c_2=50$. As Table~\ref{tab: regularized_vs_wavelet} shows, applying the action modifier forces the model to include the $0\%$ action, except for the situation that the regularizer is too strict. The modified model has improved performance for regularizer coefficients of $0.1$ and $1.0$ while changing the dose less often. For the case of $0.1$, in $62.2\%$ of the decision points over the 2,000 test patients, the decision was to keep the dose. This number is $71.9\%$ for the regularizer coefficient of $1.0$.

\input{tables/regularized_and_wavelet}

For the explainability part of the work, we chose the model with the regularizer coefficient of $0.1$, which performs better. The actions used by this model on the test patients are ${-50\%, 0.0\%, 60\%}$. We applied Python's ``DecisionTreeClassifier'' class to the maintenance portion of dosing trajectories in the test set with INR as the only input variable and the percent dose change as the label. Then we manually combined similar and redundant leaves. The final tree can fit in a simple table. Table~\ref{tab: explainable_dosing_protocol} only has three decisions.

\input{tables/explainable_protocol}

We tested all protocols on our test patients to see if this explainable protocol could compete with the baseline protocols. Since we built the explainable protocol on the test set of the DRL model, we generated a new test dataset of 2,000 patients and compared the performances. Table~\ref{tab: pttr_tabular_vs_baselines} shows that the explainable model loses performance, especially in the highly sensitive category, and all standard deviations are higher than the base model. However, it outperforms all baseline models. Our proposed explainable dosing protocol performs better than the Aurora and Intermountain dosing protocols. If we compare these protocols, it is clear that our proposed protocol has fewer decisions (only three INR ranges compared to 8 and 11 distinct decisions in Aurora and Intermountain, respectively~\cite{ravvaz2017personalized}). Two interesting observations are the cut-off values and the percentages. In both baseline maintenance dosing protocols, decision boundaries are arbitrary. For example, Aurora segments the INR values into intervals of $1.0-1.6-1.8-2.0-3.0-3.4-5.0$. Our explainable protocol sets the cut-offs inside the therapeutic range (at $2.27$ and $2.94$) to avoid out-of-range events. Regarding the action, the baselines have conservative dose changes: up to $10\%$ in Aurora and $15\%$ in Intermountain (skipping a dose and taking an immediate dose are the actions they have for extreme cases). In our case, dose changes are $50\%$ reduction and $60\%$ increase, translating into a faster response to changes in the patient's INR.

\input{tables/pttr_tabular_vs_baselines}

The promising dosing protocol that we proposed here should be considered as an example of how machine learning models can be transformed to satisfy the need for interpretable, easy to understand, and easy to use medical solutions. The power of this approach is in the fact that we do not impose any hard constraints on the model (such as the number of available decisions or the cut-off points) or incorporate direct human input into the model (such as overriding model's action during the training process). Rather, we employ a two stage approach with soft constraints and minimal guidance.

In the first stage, the DRL method is allowed to have the freedom it needs to learn the optimal policy, but incentivized through Action Forging to prefer a prolicy that has the potential to be explained. The second stage distills this policy into an explainable model in the form familiar to practitioners. Such two phase method can be applied to any sequential decision making scenario, especially in healthcare domain where strict standards need to be met.

There are still a number of issues and improvement potentials that need to be investigated. For one, we only focused on the dose change excluding the duration from the decision space. The added decision dimension (duration) increases the number of decisions dramatically, and makes the training process more challenging and data intensive. Moreover, duration needs a different form of Action Forging to persuade the model to make longer decisions whenever possible. 

Another subject that needs more in-depth study is the proposed cut-offs by the model. We believe that two factors play a role in the model's decision. First, despite inter-personal variations, any cohort of actual or virtual patients might present similarities in their response to warfarin based on the extent of diversity achieved in the dosing study. It is possible that experimenting on a different cohort result in a different set of cut-off values. Second, the reward function is the guiding signal in any RL training. Our proposed reward function penalizes out of range INR values quadratically. This might have made the model more conservative and defining the cut-off values inside the therapeutic range.

Finally, Action Forging is a useful technique that can take on many forms. We only presented two techniques, action regularizer and action focus, that were necessary in making the final model more explainable. Depending on the use case, novel forging techniques can be developed.

%% file: tables/pttr_base_vs_baselines.tex
\begin{table}[ht]
    \caption{Percent Time in Therapeutic Range of our model vs. baseline protocols}
    \label{tab: pttr_base_vs_baselines}
    \centering
    \resizebox{\columnwidth}{!}{
    \begin{tabular}{l c c c c}
        \toprule
        protocol    & Base model             & Full state    & Aurora         & Intermountain \\
        sensitivity \\
        \midrule
        normal      & \textbf{83.8\% (0.10)} & 82.9\% (0.09) & 76.4\% (0.15) & 64.7\% (0.30) \\
        sensitive   & \textbf{85.7\% (0.06)} & 84.0\% (0.08) & 60.1\% (0.26) & 46.4\% (0.34) \\
        highly sens.& \textbf{83.8\% (0.09)} & 80.9\% (0.11) & 25.9\% (0.21) & 11.1\% (0.17) \\
        \midrule
        all         & \textbf{84.5\% (0.09)} & 83.2\% (0.09) & 68.6\% (0.23) & 56.1\% (0.34) \\
        \bottomrule
    \end{tabular}
    }
\end{table}

%% file: tables/base_vs_regularized.tex
\begin{table*}[ht]
    \caption{PTTR and action count for different regularization coefficients}
    \label{tab: base_vs_regularized}
    \centering
    \begin{tabular}{l c c c c c c} 
        \toprule
        model      & Base ($0.0$)           & $10^{-4}$        & $10^{-3}$        & $0.01$        & $0.1$         & $1$          \\
        sensitivity \\
        normal            & 83.8\% (0.10)	& 84.2\% (0.10)	& 86.6\% (0.06)	& 84.3\% (0.10) & 84.6\% (0.08) & 79.4\% (0.13) \\
        sensitive         & 85.7\% (0.06)	& 86.3\% (0.06)	& 87.1\% (0.06)	& 85.2\% (0.08) & 85.4\% (0.07) & 77.2\% (0.12) \\
        highly sensitive  & 83.3\% (0.09)	& 85.9\% (0.08)	& 85.1\% (0.09)	& 79.4\% (0.11) & 81.5\% (0.11) & 63.7\% (0.15) \\
        \midrule
        all               & 84.5\% (0.09)	& 85.1\% (0.08)	& 86.7\% (0.06)	& 84.4\% (0.09) & 84.8\% (0.08) & 78.0\% (0.13) \\
        \midrule
        number of actions      & $21$       & $21$          & $21$          & $20$          & $4$           & $9$           \\
        number of actions used & $7$        & $5$           & $5$           & $4$           & $4$           & $3$           \\
        \bottomrule
    \end{tabular}
\end{table*}

%% file: tables/regularized_and_wavelet.tex
\begin{table*}[ht]
    \caption{PTTR and action count of the base model and different regularization coefficients with wavelet ($u=0.2$, $d=-0.1$, $r=1.0$)}
    \label{tab: regularized_vs_wavelet}
    \centering
    \begin{tabular}{l c c c c}
        \toprule
        model             & Base            & $0.1$         & $1.0$         & $5.0$         \\
        sensitivity \\
        normal            & 83.8\% (0.10)	& 84.4\% (0.09)	& 84.6\% (0.09) & 72.3\% (0.17)\\
        sensitive         & 85.7\% (0.06)	& 86.4\% (0.07)	& 84.6\% (0.09) & 74.8\% (0.12)\\
        highly sensitive  & 83.3\% (0.09)	& 82.0\% (0.10)	& 75.7\% (0.13) & 68.6\% (0.16)\\
        \midrule
        all               & 84.5\% (0.09)	& 85.1\% (0.08)	& 84.3\% (0.09) & 73.1\% (0.09)\\
        \midrule
        number of actions      & $21$       & $10$          & $5$           & $3$ \\
        number of actions used & $7$        & $3$           & $3$           & $2$ \\
        \% of no change decisions & $0.0\%$ & $62.2\%$    & $71.9\%$      & $0.0\%$   \\
        \bottomrule
    \end{tabular}
\end{table*}

%% file: tables/explainable_protocol.tex
\begin{table}[!ht]
    \caption{The proposed explainable dosing protocol}
    \label{tab: explainable_dosing_protocol}
    \centering
    \begin{tabular}{l r}
        \toprule
        INR Range       & Dose Change \\
        \midrule
        ~~~~~~~~~~~~$INR \leq 2.27$               & $60\%$ \\
        $2.27 < INR \leq 2.94$     & $0\%$ \\
        $2.94 < INR$               & $-50\%$ \\
        \bottomrule
    \end{tabular}
\end{table}

%% file: tables/pttr_tabular_vs_baselines.tex
\begin{table}[ht]
    \caption{PTTR and action count of the explainable model vs. baseline protocols}
    \label{tab: pttr_tabular_vs_baselines}
    \centering
    \resizebox{\columnwidth}{!}{
    \begin{tabular}{l c c c c c}
        protocol    & Base          & Explainable model & Aurora         & Intermountain \\
        sensitivity \\
        \midrule
        normal      & 83.8\% (0.10) & 84.0\% (0.06) & 76.4\% (0.15) & 65.2\% (0.30) \\
        sensitive   & 85.7\% (0.06) & 71.1\% (0.17) & 60.1\% (0.26) & 46.8\% (0.34) \\
        highly sens.& 83.8\% (0.09) & 49.3\% (0.19) & 27.0\% (0.21) & 11.5\% (0.17) \\
        \midrule
        all         & 84.5\% (0.09) & 78.0\% (0.14) & 68.8\% (0.23) & 56.6\% (0.34) \\
        \midrule
        Possible Actions & 21   & 3 & 8 & 11 \\
        \bottomrule
    \end{tabular}
    }
\end{table}

%% file: 06_conclusion.tex
In this paper, we proposed a maintenance dosing algorithm for warfarin. We modeled the problem as an MDP with percent change as action and used PPO to solve it. Then we used action modifiers (regularizer and action focus) to make the action space sparser with more focus on keeping the dose unchanged. Finally, we applied a decision tree to dosing data and showed that the final model could be simpler and more effective than current dosing protocols.